\documentclass{article}
\usepackage{spconf,amsmath,graphicx,hyperref}
\usepackage{bm}
\usepackage{amssymb}
\usepackage{multirow}
\usepackage{xcolor}
\usepackage{booktabs}
\usepackage{tabularx}
\newcolumntype{Y}{>{\centering\arraybackslash}X}

\usepackage[normalem]{ulem}

\definecolor{cvprblue}{rgb}{0.21,0.49,0.74}
\hypersetup{colorlinks=true,
            allcolors=cvprblue,
            breaklinks=true,
            pdfborder={0 0 0}}

\title{Latent Space Is Not Flat:\\
Rethinking Latent Structure for 3D Medical Image Synthesis}

\name{
\begin{tabular}{c}
Haowen Xue$^{1}$ \quad
Hao Chen$^{2}$ \quad
Hexuan Hu$^{1}$ \quad
Qian Huang$^{1,\dagger}$ \quad
Yi Han$^{1}$ \\[2pt]
Qing Meng$^{1}$ \quad
Zaipeng Xie$^{1}$ \quad
Chao Li$^{2}$ \quad
Haoli Xu$^{3}$
\end{tabular}
\thanks{$\dagger$~Corresponding author.}
}
 
\address{\quad  $^{1}$Hohai University \quad$^{2}$University of Cambridge  \quad $^{3}$National University of Defense Technology}

\begin{document}
\ninept

\maketitle
\suppressfloats[t]

\begin{abstract}

Latent generative models make 3D medical image synthesis computationally practical by generating in a compressed space. However, we show that the common flat Euclidean assumption induced by $\ell_2$ objectives is imprecise: latent-space geometry is so strongly anisotropic that equal-magnitude errors can produce drastically different decoded distortions.  
We further find that this anisotropy has a clear feature: sensitive variation concentrates in a low-rank subspace. The dominant low-rank components capture the overall \textit{structure}, encoding long-range, spatially coordinated variation while remaining resistant to local noise. Its orthogonal \emph{residual}, in contrast, mainly captures local and image-specific variation.
Motivated by this asymmetry, we introduce Latent Structure Flow (LSF). At each block, LSF decomposes the latent state into structure and residual, models structural changes with global context, and predicts residual variation locally while preserving a direct path for the input structure. LSF changes only the generator, leaving the frozen codec and pointwise training objective unchanged. Across cross-modality synthesis and tumor inpainting tasks, LSF outperforms all compared baselines on both global and tumor-specific metrics, demonstrating the benefit of explicitly modeling latent-space structure for 3D medical image synthesis.

\end{abstract}


\begin{figure}[t]
\centering
\includegraphics[width=\columnwidth]{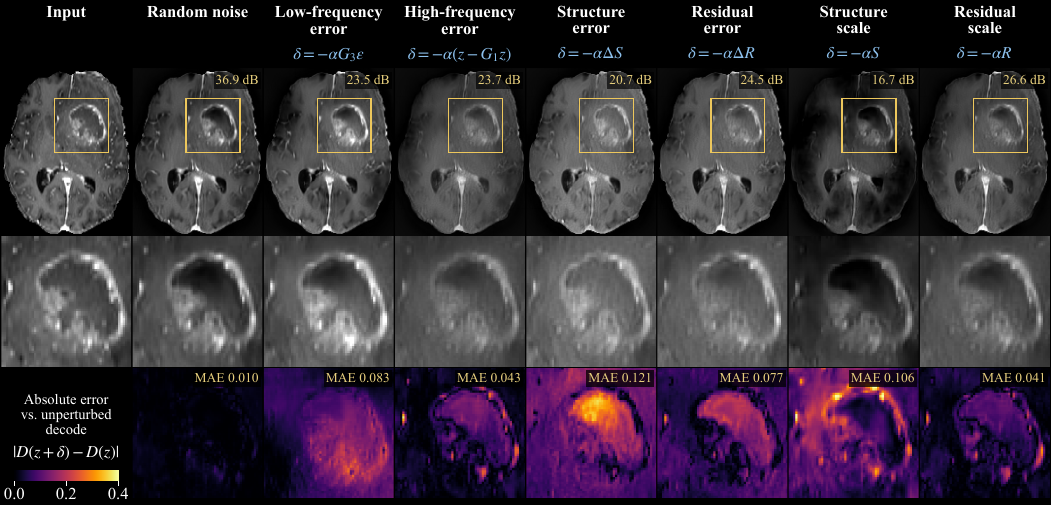}
\vspace{-10pt}
\caption{Decoded effect of latent errors of equal norm on one subject: slice with PSNR, lesion crop, and absolute error with MAE. Attenuating the structure causes the largest distortion; residual errors change only local appearance.}
  \label{fig:teaser}
\vspace{-12pt}
\end{figure}

% Figure 2: framework. Input in main.tex before the introduction so it lands on page 2.
\begin{figure*}[t]
\centerline{\includegraphics[width=\textwidth]{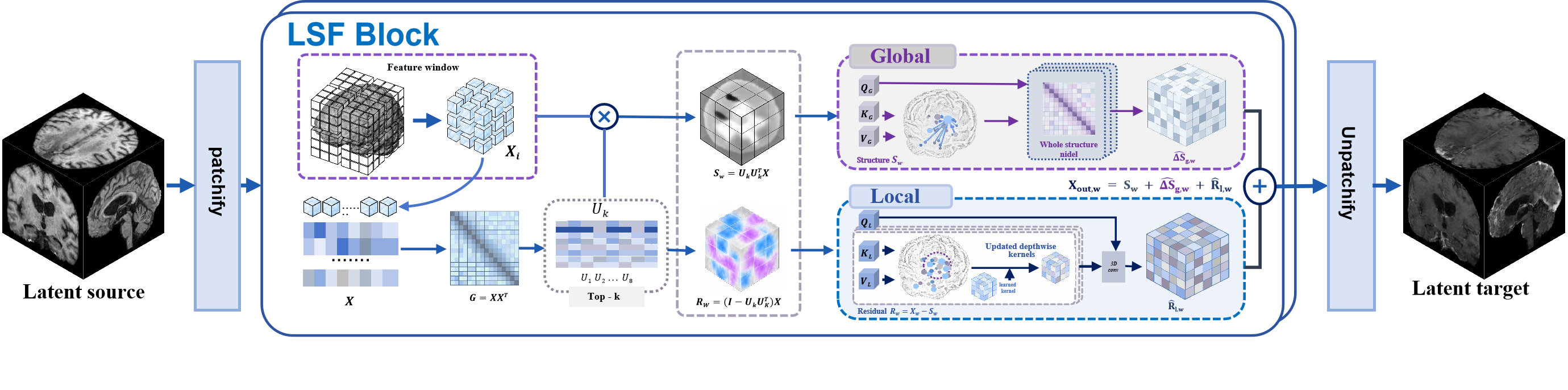}}
\caption{\textbf{Latent Structure Flow.} In every block, a basis estimated jointly from the state and the condition splits the input tokens into structure $S_w$ and residual $R_w$ (Eq.~\eqref{eq:basis}). $S_w$ is routed to the global branch and $R_w$ to the local branch; the output adds the predicted structural change and output residual to $S_w$ (Eq.~\eqref{eq:inherit}).
}
\label{fig:overview}
\end{figure*}
% Table 1 moved to page 3: input at the start of 03_method.tex.

\section{Introduction}
\label{sec:intro}

Latent generative modeling reduces the computational cost of synthesizing 3D medical images by operating on compressed representations~\cite{rombach2022latent,guo2025maisi,khader2023medical}. The generator predicts a target latent and is trained by measuring error directly in latent space.

However, the loss measures how large the latent error is, but not where it lies. Prior work has observed that small latent errors can still decode poorly~\cite{berrada2024lpl,morae2026,crossflow2026}, that latent components contribute unequally to the decoded image~\cite{sphericalflow2026,rao2026inversion}, and, in medical imaging, that reconstruction fidelity does not predict how learnable a latent is~\cite{learnabilitygap2026,tokgen2026}. A latent generator can achieve low pointwise error while still producing visibly degraded reconstructions, including blurred details and faded lesion contrast.  To locate where this failure arises, we probe three frozen autoencoders with errors of fixed latent norm, varying only how the error is organized, from random noise to content-aligned errors. %The analysis shows that the decoder is resistant to random perturbations but strongly amplifies more spatially coherent errors; Gaussian noise and structure attenuation differ by 19.6\,dB in decoded PSNR on MAISI
At matched latent error norms, attenuating the structural component causes greater decoded distortion than adding noise within that component on MAISI (Table~\ref{tab:sr}A).   

We then ask whether these errors share a common representation in the latent. They do. Within each local window, we decompose the latent $z$ into two components. The dominant eigenvectors of the window's Gram matrix describe the spatial patterns that carry the largest share of its energy; with $P$ the projection onto them, we call $S=Pz$ the \emph{structure} and its orthogonal complement $R=(I-P)z$ the \emph{residual}. %For an over-smoothed error, 58--80\% of the error energy falls in the \emph{structure}, against 33--34\% for Gaussian noise. Replacing the structural error with noise in the residual improves decoded PSNR by 4.0--6.9\,dB, whereas the reverse improves it by only 0.6--2.0\,dB (Table~\ref{tab:sr}); Fig.~\ref{fig:teaser} illustrates the effect on one subject. 
Across the three tested codecs, 58–80\% of the over-smoothed error energy lies in the structural component  (Table~\ref{tab:sr}). Together with the equal-norm perturbation results, this observation motivates modeling structural and residual variation separately. Fig.~\ref{fig:teaser}  illustrates the decoded effects of different perturbations on one subject.
Wavelet-based methods~\cite{friedrich2024cwdm} also decompose the representation locally, but their split is based on scale, whereas ours is based on energy.

This raises a question: how can a generator get the structure right? First, the structure is largely shared between modalities. In the translation task, the structural subspaces of the input and output modalities overlap by 0.66--0.87, whereas their residuals are only weakly correlated (cosine 0.18--0.25, against 0.02 across subjects). 
Second, the two components exhibit correlations over different spatial ranges. 
The structure stays correlated over long distances (cosine 0.21--0.42 at about half the brain width), while the residual is nearly uncorrelated at the same distance (at most 0.04). 
These statistics motivate global processing for structure and local processing for residual.

We build \emph{Latent Structure Flow} (LSF) on these two properties. LSF is a flow model whose token mixer has a global and a local branch. 
In each block, it splits the tokens, which carry both the noisy state and the condition, into structure and residual, and routes the structure to the global branch and the residual to the local branch; 
the resulting structural correction and full output residual are added to the input-token structure to form the output tokens. Unlike objectives that add perceptual terms through the decoder, which is costly for 3D volumes~\cite{berrada2024lpl,pcflow2026}, or through a learned latent network~\cite{kang2024elatentlpips}, and unlike methods that adapt the codec to the generator~\cite{sphericalflow2026,morae2026}, LSF changes only the generator: the autoencoder, the pointwise flow-matching loss, and the sampler stay the same.

%Our contributions are an analysis locating damaging latent errors in a content-defined structure, and LSF, which uses this decomposition to guide both feature processing and target prediction. We evaluate LSF on modality translation and tumor inpainting. Against the matched baseline (Table~\ref{tab:ablation}), LSF gains 0.85\,dB in tumor PSNR and lowers the share of its own latent error that falls in the structure from 11.5\% to 8.0\%. Removing the structural skip costs 0.6\,dB, removing the split as well 0.85\,dB, and reversing the routing 1.2\,dB.

Our contributions are a controlled analysis of structure and residual components in frozen medical latents, and LSF, which uses this decomposition for global/local routing and structural skip connections. Experiments on cross-modality synthesis and tumor inpainting evaluate the resulting gains against external baselines and matched ablations. The project repository is hosted at \url{https://github.com/redcake4/LDM}.

% 02_analysis merged back into the introduction (v8)
\section{Method}
\label{sec:method}

\label{sec:method_setup}

\noindent\textbf{Setup.} Let $x_t$ be a target medical volume and $c$ an aligned image condition. A frozen autoencoder maps them to $z_t=E(x_t)$ and $z_c=E(c)$ on a common latent grid, with $z_t\in\mathbb R^{C\times d\times h\times w}$. The generator predicts $\widehat z_t$, which is decoded as $D(\widehat z_t)$.

For a prediction error $\delta=\widehat z_t-z_t$, an $\ell_2$ loss measures only $\|\delta\|_2^2$, whereas the decoded damage depends on where $\delta$ lies (Sec.~\ref{sec:intro}). LSF keeps this loss and changes only how $\widehat z_t$ is produced.

\label{sec:method_structure}
\label{sec:analysis}

\subsection{Latent Decomposition} We partition a latent into non-overlapping $p^3$ patches and group the resulting tokens into windows. A window is a matrix $X_w\in\mathbb R^{K\times p^3C}$ with $K$ token positions. Let $U_w\in\mathbb R^{K\times r}$ contain the top-$r$ eigenvectors of $X_wX_w^{\top}$. The projector $P_w=U_wU_w^{\top}$ defines
\begin{equation}
S_w=P_wX_w,\qquad R_w=(I-P_w)X_w.
\label{eq:sr}
\end{equation}
We call $S_w$ the structure and $R_w$ the residual. $P_w$ depends only on the latent, not on the decoder. Gaussian error places 33--34\% of its energy in the structure (Table~\ref{tab:sr}).

An error splits the same way, $\|\delta\|^2=\|P\delta\|^2+\|(I-P)\delta\|^2$, so the pointwise loss is blind to the split. The structural fraction $\rho_S$ (Eq.~\eqref{eq:rho}) measures this split.

% Fig. 2 (fig:overview) is in figures/framework.tex, input in main.tex so it lands on page 2.

\label{sec:method_generator}

\subsection{Latent Generation}
LSF applies the decomposition within each block to guide feature routing and token updates.
The noisy state and the condition are concatenated, $[z_\tau\Vert z_c]$, patch-embedded into $N$ tokens and processed by $L$ blocks; 
an output readout maps the final tokens to the clean target $\widehat z_t$~\cite{li2025jit}
(Fig.~\ref{fig:overview}). Each block has an adaLN-modulated RMSNorm, a token mixer and a SwiGLU feed-forward layer~\cite{peebles2023dit}. 
In each block, the normalized mixer input
is split into $S$ and $R$ (Eq.~\eqref{eq:basis}). The two parts pass through the global and local branches, and their gated outputs are fused (Eq.~\eqref{eq:proj}). The result is added to the residual stream with an adaLN gate, followed by the SwiGLU layer. 

\noindent\textbf{Global and local branches.} 
The mixer has two sample-adaptive branches, using inner-update operators inspired by prior work~\cite{sun2024ttt,han2026vit3}.
A branch projects its input to queries, keys and values, takes one normalized gradient step on an inner model $f_\theta$ from a learned initialization $\theta_0$,
\begin{equation}
\theta=\theta_0-\lambda\,\frac{\nabla\ell(\theta_0)}{\|\nabla\ell(\theta_0)\|+1},\qquad \ell(\theta)=-\frac{1}{N}\sum_{i}\langle f_\theta(k_i),v_i\rangle,
\label{eq:inner}
\end{equation}
and returns $f_\theta(q_i)$ for every query. In the global branch, $f_\theta(q)=(q\Theta_1)\odot\operatorname{SiLU}(q\Theta_2)$ is a per-head gated map fitted over the whole volume, so every query sees global context. In the local branch, $f_\theta(q)=\theta*q$ is a depthwise $3{\times}3{\times}3$ convolution, applied at dilations $\{1,2,4\}$ and averaged, 
so each readout aggregates its neighbourhood using a kernel adapted to the whole volume.

\noindent\textbf{Feature routing.}
Let $H\in\mathbb R^{N\times d}$ denote the normalized mixer input. We split it window by window with Eq.~\eqref{eq:sr}, $H=S+R$, using a basis estimated jointly from $H$ and the condition features $C$ of a separate patch embedding:
\begin{equation}
\begin{aligned}
\Gamma(A)&=\frac{AA^{\top}}{\operatorname{tr}(AA^{\top})+\eta},\\
U_w&=\operatorname{eig}_r\!\big(\Gamma(C_w)+\Gamma(H_w)\big),\quad P_w=U_wU_w^{\top},
\end{aligned}
\label{eq:basis}
\end{equation}
where $\operatorname{eig}_r$ returns the top-$r$ eigenvectors. Both Gram matrices are $K\times K$, so they can be added although $C_w$ and $H_w$ have different features, and trace normalization keeps either from dominating by scale. The condition indicates where the structure lies, while the state lets the basis adapt during sampling. The basis is detached. The structure feeds the global branch $\mathcal G$ and the residual the local branch $\mathcal D$:
\begin{equation}
\begin{aligned}
(Q_g,K_g,V_g)&=SW_g,\qquad (Q_l,K_l,V_l)=RW_l,\\
Y&=\operatorname{Proj}\big([\,g_g\odot\mathcal G(S)\,\Vert\, g_l\odot\mathcal D(R)\,]\big).
\end{aligned}
\label{eq:proj}
\end{equation}
% Table 1 (input from 04_experiments.tex, Sec. Latent analysis, so it lands on page 3): structure S vs residual R within each frozen codec. Data: ~/Documents/OverleafWork/ICASSP2027_SAGFlow-3D (latent_probe, probe_scale). Previous wide layout archived in OverleafWork/tables/.
\begin{table}[t]
\caption{\textbf{Latent $S$ vs.\ $R$  error modes}. Perturbations have equal latent norm. (A) Attenuating $S$ hurts most. (B) Over-smoothed error falls mostly in $S$; Gaussian error does not. (C) $S$ is shared within and across patients, $R$ is not. (D) $S$ is long-range, $R$ is local. (E) One global basis captures $S$ but not $R$.}
\label{tab:sr}
\label{tab:sr_properties}
\centering\scriptsize
\setlength{\tabcolsep}{4pt}
\setlength{\aboverulesep}{0pt}\setlength{\belowrulesep}{0pt}\renewcommand{\arraystretch}{1.3}
\resizebox{\columnwidth}{!}{\begin{tabular}{@{}l|cc|cc|cc@{}}
\toprule
\multirow{2}{*}{Property} & \multicolumn{2}{c|}{MAISI} & \multicolumn{2}{c|}{KL-VAE} & \multicolumn{2}{c}{VQ-VAE} \\
 & $S$ & $R$ & $S$ & $R$ & $S$ & $R$ \\
\midrule
(A) Attenuation PSNR (dB) & $24.6_{\pm 0.9}$ & $33.1_{\pm 1.3}$ & $31.8_{\pm 2.3}$ & $36.3_{\pm 1.2}$ & $29.6_{\pm 2.3}$ & $29.5_{\pm 1.8}$ \\
(A) Noise PSNR (dB)       & $35.4_{\pm 1.0}$ & $30.5_{\pm 1.4}$ & $37.9_{\pm 0.9}$ & $33.4_{\pm 1.6}$ & $33.6_{\pm 1.9}$ & $30.5_{\pm 2.1}$ \\
\midrule
(B) Smooth error (\%)   & $58_{\pm 0.8}$ & $42_{\pm 0.8}$ & $80_{\pm 1.5}$ & $20_{\pm 1.5}$ & $76_{\pm 4.6}$ & $24_{\pm 4.6}$ \\
(B) Gaussian error (\%) & $34_{\pm 0.1}$ & $66_{\pm 0.1}$ & $34_{\pm 0.3}$ & $66_{\pm 0.3}$ & $33_{\pm 0.3}$ & $67_{\pm 0.3}$ \\
\midrule
(C) Within patient  & $0.87_{\pm 0.01}$ & $0.25_{\pm 0.06}$ & $0.66_{\pm 0.02}$ & $0.18_{\pm 0.07}$ & $0.84_{\pm 0.01}$ & $0.22_{\pm 0.12}$ \\
(C) Across patients & $0.77_{\pm 0.03}$ & $0.02_{\pm 0.01}$ & $0.59_{\pm 0.02}$ & $0.02_{\pm 0.02}$ & $0.76_{\pm 0.03}$ & $0.03_{\pm 0.02}$ \\
\midrule
(D) Long-range corr. & $0.40_{\pm 0.03}$ & $0.04_{\pm 0.01}$ & $0.21_{\pm 0.02}$ & $0.01_{\pm 0.00}$ & $0.42_{\pm 0.02}$ & $0.03_{\pm 0.01}$ \\
\midrule
(E) Shared basis (\%) & $57_{\pm 4}$ & $10_{\pm 1}$ & $65_{\pm 2}$ & $26_{\pm 1}$ & $95_{\pm 1}$ & $14_{\pm 1}$ \\
\bottomrule
\end{tabular}}
\end{table}
% Table 1, single column; placed here so the [t] float lands at the top of page 3
% figures/latent_SR.tex -- \input from 04_experiments.tex (label fig:latent_sr, do not rename).
% PDF drawn at true \columnwidth: 252 pt x 76.8 pt, 7 panels of 32.3 pt, 5.3 pt headers (no blue sub-header).
% Source: /Users/haochen/repo/J3D/figures/latent_SR_teaser/make_SR_teaser.py
%   -> SR_rank_2case_20260924-0940.pdf
\begin{figure}[t]
\centering
\includegraphics[width=\columnwidth]{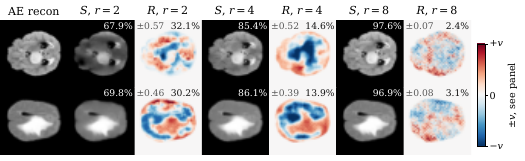}
\vspace{-12pt}
%\caption{\textbf{%Choosing the rank.} $S$ and $R$ decoded separately for $r{\in}\{2,4,8\}$; each
%panel gives its share of the latent energy. Beyond $r{=}2$ the residual is drained to a few
%percent, so we use $r{=}2$.
\caption{\textbf{Choosing the rank.} $S$ and $R$ are decoded separately
for $r\in\{2,4,8\}$; each panel gives its share of the latent
energy. The residual retains 30--32\% of the latent energy
at $r{=}2$, but only 2--3\% at $r{=}8$, motivating our choice of $r{=}2$.}
\label{fig:latent_sr}
\vspace{-12pt}
\end{figure}
% Fig., single column; follows Table 1 so both [t] floats land on page 3
% \input{figures/latent_SR_row}  % single-row variant of the same panels; headers are ~3 pt
 The assignment matches each operator to the statistics of its input. The global branch fits a single map for the whole volume, which suits content that shares one global basis: such a basis of dimension 8 explains 57--95\% of the structure energy but only 10--26\% of the residual energy (Table~\ref{tab:sr}). The local branch adapts a small kernel that each position applies to its neighbourhood, which suits the residual, 
whose long-range correlation is weak.
Row (e) of Table~\ref{tab:ablation} reverses the assignment.

\noindent\textbf{Intermediate feature fusion.}
Within each token mixer, the two branch feature outputs are rescaled by token-wise gates
$g_g,g_l\in(0,2)$, initialized to one and predicted by a small MLP from $S$, $\|S\|^2$, $\|R\|^2$ and the time embedding; they are then concatenated, projected back to $d$ channels (Eq.~\eqref{eq:proj}) and normalized by a time-modulated RMSNorm. 

\noindent\textbf{Output-token composition.}\space
Let $X_{\mathrm{in},w}=H_w$ denote the input tokens to the mixer in window $w$. Their structure is $S_w=P_wX_{\mathrm{in},w}$, using the same projector as the feature routing in Eq.~\eqref{eq:basis}. %Denote the global branch's structural-change prediction by $\widehat{\Delta S}_{g,w}$ and the local branch's full output-residual prediction by $\widehat R_{l,w}$. 
For each window, the global branch predicts a structural correction  $\widehat{\Delta S}_{g,w}$, while the local branch predicts the full output residual $\widehat R_{l,w}$. Both predictions are mapped to the input-token dimension through their respective output projections .
The output tokens are
\begin{equation}
X_{\mathrm{out},w}=S_w+\widehat{\Delta S}_{g,w}+\widehat R_{l,w}.
\label{eq:inherit}
\end{equation}

%The skip connection carries only the input structure $S_w$, leaving the residual to be generated anew.
This structural shortcut replaces the conventional full-input mixer shortcut, so the input residual is not directly carried into the output. 

\label{sec:method_training}
\label{sec:method_sampling}

\noindent\textbf{Training and sampling.} We train with flow matching~\cite{lipman2023flow}: with $z_\tau=(1-\tau)\epsilon+\tau z_t$ and $\epsilon\sim\mathcal N(0,I)$, the loss is $\mathcal L=\mathbb E\,\|\widehat z_t-z_t\|_2^2/a_\tau^2$ with $a_\tau=\max(1-\tau,\varepsilon)$, which is velocity matching away from the floor $\varepsilon$. No component-specific weighting or decoder gradient is used.  An Euler solver integrates $(\widehat z_t-z_\tau)/a_\tau$ from noise, and the final latent is decoded once. The split is applied in every block. For inpainting, the hole mask is appended to $z_c$ as an extra channel. In fully masked windows $\Gamma(C_w)=0$, so the basis comes from the mixer input alone; its structural component remains the reference for the token update. Known voxels are copied back after decoding. 
The split adds one $K{\times}K$ eigendecomposition per window in every block; the token update reuses this projector.
% Tables 2-3 (full width) are input here so that they float to the top of page 4.
% Table 2: cross-contrast translation.
\begin{table*}[t]
\caption{\textbf{Cross-contrast translation} (mean$_{\pm\mathrm{SD}}$; best in bold). MAE, PSNR and SSIM over the brain ($\times10^{2}$ for MAE); Tumor and Change: MAE in the tumor and in the change region ($\times10$). Baselines operate in pixel space.}
\label{tab:main}
\centering\scriptsize
\setlength{\tabcolsep}{1pt}
\setlength{\aboverulesep}{0pt}\setlength{\belowrulesep}{0pt}\renewcommand{\arraystretch}{1.3}
\begin{tabularx}{\textwidth}{@{}l|YYYYY|YYYYY@{}}
\toprule
 & \multicolumn{5}{c|}{\textbf{T1n$\to$T1c}} & \multicolumn{5}{c}{\textbf{T2w$\to$T2f}} \\
\midrule
Method & MAE$\downarrow$ & PSNR$\uparrow$ & SSIM$\uparrow$ & Tumor$\downarrow$ & Change$\downarrow$ & MAE$\downarrow$ & PSNR$\uparrow$ & SSIM$\uparrow$ & Tumor$\downarrow$ & Change$\downarrow$ \\
\midrule
cWDM~\cite{friedrich2024cwdm}    & $1.59_{\pm 0.60}$ & $26.21_{\pm 2.21}$ & $0.921_{\pm 0.02}$ & $1.365_{\pm 0.50}$ & $1.410_{\pm 0.72}$ & $1.57_{\pm 0.46}$ & $26.14_{\pm 1.84}$ &$0.923_{\pm 0.01}$ & $1.348_{\pm 0.48}$ & $1.451_{\pm 0.50}$\\
PMRF~\cite{brandstotter2025pmrf} & $1.64_{\pm 0.63}$ & $26.14_{\pm 2.31}$ & $0.918_{\pm 0.01}$ & $1.376_{\pm 0.55}$ & $1.616_{\pm 0.82}$ & $1.59_{\pm 0.43}$ & $26.18_{\pm 1.80}$ & $0.916_{\pm 0.01}$ & $1.359_{\pm 0.45}$ & $1.506_{\pm 0.48}$\\
MoTFM~\cite{yazdani2025motfm}    & $2.06_{\pm 0.64}$ & $24.72_{\pm 2.73}$ & $0.901_{\pm 0.07}$ & $1.851_{\pm 0.72}$ & $1.699_{\pm 0.82}$ & $1.70_{\pm 0.54}$  & $25.05_{\pm 1.88}$ & $0.904_{\pm 0.02}$ & $1.354_{\pm 0.45}$ & $1.566_{\pm 0.54}$ \\
WFM~\cite{tur2026wfm}            & $1.88_{\pm 0.65}$ & $24.86_{\pm 2.42}$ & $0.904_{\pm 0.03}$ & $1.393_{\pm 0.63}$ & $1.605_{\pm 0.82}$ & $1.63_{\pm 0.49}$ & $25.12_{\pm 1.64}$ & $0.908_{\pm 0.02}$ & $1.383_{\pm 0.47}$ & $1.522_{\pm 0.53}$\\
\midrule
LSF (ours)                 & $\mathbf{1.56}_{\pm 0.59}$ & $\mathbf{26.29}_{\pm 2.15}$ & $\mathbf{0.927}_{\pm 0.01}$ & $\mathbf{1.341}_{\pm 0.59}$ & $\mathbf{1.357}_{\pm 0.71}$ & $\mathbf{1.54}_{\pm 0.40}$ & $\mathbf{26.44}_{\pm 1.92}$ & $\mathbf{0.939}_{\pm 0.01}$ & $\mathbf{1.324}_{\pm 0.45}$ & $\mathbf{1.424}_{\pm 0.43}$ \\
\bottomrule
\end{tabularx}
\end{table*}

% Table 3: tumor inpainting.
\begin{table*}[t]
\caption{\textbf{Tumor inpainting} (mean$_{\pm\mathrm{SD}}$; best in bold). PSNR (dB) and SSIM inside the whole-tumor region. }
\label{tab:inpaint}
\centering\scriptsize
\setlength{\tabcolsep}{1pt}
\setlength{\aboverulesep}{0pt}\setlength{\belowrulesep}{0pt}\renewcommand{\arraystretch}{1.3}
\begin{tabularx}{\textwidth}{@{}l|YY|YY|YY|YY|YY@{}}
\toprule
 & \multicolumn{2}{c|}{\textbf{T1c}} & \multicolumn{2}{c|}{\textbf{T1n}} & \multicolumn{2}{c|}{\textbf{T2f}} & \multicolumn{2}{c|}{\textbf{T2w}} & \multicolumn{2}{c}{\textbf{Average}} \\
\midrule
Method & PSNR$\uparrow$ & SSIM$\uparrow$ & PSNR$\uparrow$ & SSIM$\uparrow$ & PSNR$\uparrow$ & SSIM$\uparrow$ & PSNR$\uparrow$ & SSIM$\uparrow$ & PSNR$\uparrow$ & SSIM$\uparrow$ \\
\midrule
LeFusion~\cite{zhang2024lefusion} & $24.04_{\pm 3.52}$ & $0.702_{\pm 0.09}$ & $25.50_{\pm 3.75}$ & $0.695_{\pm 0.11}$ & $18.35_{\pm 4.22}$ & $0.547_{\pm 0.12}$ & $19.96_{\pm 4.30}$ & $0.590_{\pm 0.11}$ & $21.96$ & $0.633$ \\
DiffTumor~\cite{chen2024towards} & $25.59_{\pm 2.73}$ & $0.738_{\pm 0.07}$ & $26.04_{\pm 3.44}$ & $0.704_{\pm 0.10}$ & $21.12_{\pm 2.85}$ & $0.608_{\pm 0.09}$ & $22.05_{\pm 3.03}$ & $0.636_{\pm 0.08}$ & $23.70$ & $0.671$ \\
cWDM~\cite{friedrich2024cwdm} & $26.33_{\pm 2.83}$ & $0.774_{\pm 0.06}$ & $26.28_{\pm 3.14}$ & $0.711_{\pm 0.09}$ & $20.77_{\pm 3.29}$ & $0.598_{\pm 0.10}$ & $22.01_{\pm 2.70}$ & $0.636_{\pm 0.08}$ & $23.85$ & $0.680$ \\
\midrule
LSF (ours) & $\mathbf{26.53}_{\pm 2.57}$ & $\mathbf{0.796}_{\pm 0.07}$ & $\mathbf{27.00}_{\pm 3.26}$ & $\mathbf{0.728}_{\pm 0.10}$ & $\mathbf{22.91}_{\pm 2.90}$ & $\mathbf{0.672}_{\pm 0.08}$ & $\mathbf{24.10}_{\pm 2.66}$ & $\mathbf{0.720}_{\pm 0.08}$ & $\mathbf{25.14}$ & $\mathbf{0.729}$ \\
\bottomrule
\end{tabularx}
\end{table*}

\begin{figure}[t]
\centering
\includegraphics[width=\columnwidth]{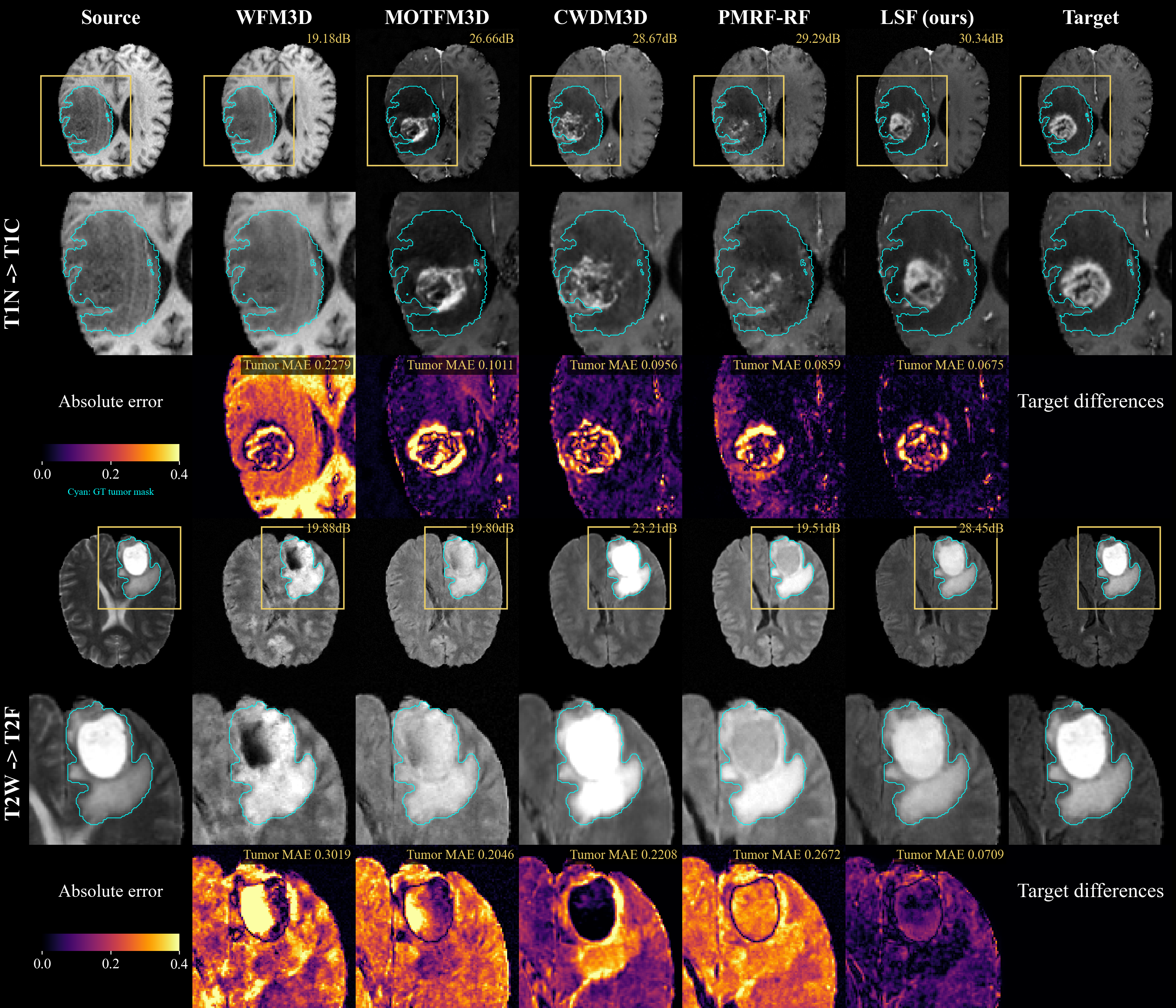}
\vspace{-10pt}
\caption{Cross-contrast translation cases, T1n$\to$T1c (top) and T2w$\to$T2f (bottom): slice, tumor-centered crop (yellow box) and absolute error, with the ground-truth tumor contour in cyan. Annotations give full-volume PSNR and tumor MAE.}
\label{fig:translation_qual}
\vspace{-12pt}
\end{figure}
\section{Experiments}
\label{sec:exp}

% Table 4: ablation.
\begin{table}[t]
\caption{\textbf{Ablation} on tumor inpainting. All metrics are computed in the tumor region; LPIPS $\times10^{-3}$; $\rho_S$ in \% with $r{=}2$ (Gaussian error: 6.25\%). cWDM is the strongest pixel-space baseline.}
\label{tab:ablation}
\centering\footnotesize
\setlength{\tabcolsep}{3pt}
\setlength{\aboverulesep}{0pt}\setlength{\belowrulesep}{0pt}\renewcommand{\arraystretch}{1.25}
\begin{tabular*}{\columnwidth}{@{}c@{\hspace{6pt}}l@{\hspace{6pt}}|cccc@{}}
\toprule
 & Configuration & PSNR$\uparrow$ & SSIM$\uparrow$ & LPIPS$\downarrow$ & $\rho_S\downarrow$ \\
\midrule
(a) & Global branch only & 23.27 & 0.672 & 1.55 & 12.6 \\
(b) & + local branch & 24.29 & 0.703 & 1.52 & 11.5 \\
(c) & + split: $S\!\to$global, $R\!\to$local & 24.51 & 0.712 & 1.50 & 8.7 \\
(d) & + input-token structure $S$ (LSF) & 25.14 & 0.729 & 1.44 & 8.0 \\
\midrule
(e) & (d) with $S\!\to$local, $R\!\to$global & 23.90 & 0.697 & 1.60 & 10.9 \\
\midrule
(f) & cWDM~\cite{friedrich2024cwdm}, pixel space & 23.85 & 0.680 & 1.48 & 11.8 \\
\bottomrule
\end{tabular*}
\end{table}

% \subsection{Setup}
% \label{sec:exp_setup}

\noindent\textbf{Data.} We use BraTS-GLI~\cite{baid2021brats}, split 8:1:1 into training, validation and test sets. Volumes are 1\,mm isotropic. The frozen MAISI autoencoder~\cite{guo2025maisi} is our codec.

\noindent\textbf{Implementation.} LSF has 10 blocks with hidden size 512. The global branch uses 8 heads; the local branch uses a 64-channel depthwise $3{\times}3{\times}3$ kernel at dilations $\{1,2,4\}$; both take one inner step with $\lambda{=}1$. The codec downsamples by $4\times$ and latents are patch-embedded with $p{=}2$, giving a $24{\times}32{\times}32$ token grid; windows hold $2{\times}4{\times}4$ tokens, the basis has rank $r{=}2$, and the projector is detached from the gradient. The model has 62.9M parameters and one forward pass costs 1.72\,TFLOPs. We train with AdamW ($\beta_1{=}0.9$, $\beta_2{=}0.95$), a constant learning rate of $3{\times}10^{-5}$ after 2 warm-up epochs, no weight decay, one volume per batch, bf16 autocast and an EMA of 0.9999, for 80 epochs on one RTX A6000 (about 7\,min per epoch). The checkpoint is selected by validation MAE inside the hole. Sampling uses 50 Euler steps.

\noindent\textbf{Tasks.} \emph{Translation} synthesizes one contrast from another, T1n$\to$T1c and T2w$\to$T2f. \emph{Tumor inpainting} synthesizes tumor tissue inside a mask from the surrounding image, for each of the four contrasts.

\noindent\textbf{Baselines and metrics.} 
We compare with models
for translation~\cite{friedrich2024cwdm,brandstotter2025pmrf,yazdani2025motfm,tur2026wfm} and inpainting~\cite{friedrich2024cwdm,zhang2024lefusion,chen2024towards}, 
trained on the same split with their released settings. %TumorFlow also operates in latent space.
We report MAE, PSNR and SSIM in the brain, MAE in the tumor and in the region that changes between contrasts. For the ablation, PSNR, SSIM, LPIPS and the structural fraction $\rho_S$ are computed in the tumor region; SSIM and LPIPS are computed on 3D volumes, the latter with MedicalNet ResNet-10 features~\cite{chen2019med3d}.
The structural fraction is the share of a model's latent error $\delta=\widehat z_t-z_t$ that falls in the structure of the target,
\begin{equation}
\rho_S=\frac{\sum_w\|P^{\star}_w\delta_w\|_F^2}{\sum_w\|\delta_w\|_F^2},
\label{eq:rho}
\end{equation}
where $\delta_w$ is the error in window $w$ and $P^{\star}_w$ is the projector of Eq.~\eqref{eq:sr} computed from the target latent $z_t$. Since $\|\delta\|^2=\sum_w\|P^{\star}_w\delta_w\|^2+\|(I-P^{\star}_w)\delta_w\|^2$, $\rho_S$ is between 0 and 1; Gaussian error gives 0.33--0.34 (Table~\ref{tab:sr}), and a lower value means the remaining error lies in the residual, which the decoder tolerates better. It is a diagnostic and never enters training or model selection.

\subsection{Latent analysis}
\label{sec:exp_probe}

Table~\ref{tab:sr} probes three frozen codecs, MAISI~\cite{guo2025maisi}, a KL-VAE~\cite{lozupone2025ldae} and a VQ-VAE~\cite{prima2025}, on ten GLI volumes (five subjects, two contrast pairs). Every perturbation has the same latent norm, $\|\delta\|=0.1\|z\|$, and PSNR is measured against the unperturbed decode $D(z)$, so reconstruction error is excluded. $S$ and $R$ use rank 8 in windows of $2{\times}4{\times}4$ tokens (16\,mm per token). Fig.~\ref{fig:latent_sr} decodes the two components on their own and reports the energy each carries: at $r{=}8$ the structure already holds $97\%$ of the latent energy, %so the split is only legible at lower rank,
whereas at $r{=}2$  the residual retains  $30\text{--}32\%$. Three findings follow. 
First, attenuating $S$ causes more damage than adding noise to $S$,
and over-smoothed errors fall mostly in $S$ (A, B). Second, $S$ is shared across contrasts and subjects while $R$ is not (C), 
which motivates estimating the token decomposition jointly from the input features and the condition.
Third, $S$ is correlated over long ranges and is captured by one global basis, while $R$ is local (D, E), which motivates the global/local routing. Fig.~\ref{fig:teaser} shows these error modes on one subject.

\subsection{Main results}
\label{sec:exp_main}

\noindent\textbf{Translation.} 
%On T1n$\to$T1c (Table~\ref{tab:main}), LSF has the lowest error of all methods in the brain, in the tumor and in the region that changes between contrasts, and the highest PSNR (26.29\,dB). The largest margin is in the change region, 3.8\% below the best pixel-space model, which is where the target differs most from the condition. On T2w$\to$T2f, LSF improves on MoTFM in brain MAE (1.54 vs.\ 1.70) and PSNR (+1.4\,dB) and in the tumor (1.324 vs.\ 1.354). LSF also has the lowest change-region MAE (1.424 vs.\ 1.451 for the next-best method).
%LSF obtains these results in the compressed latent space and decodes once.
%Fig.~\ref{fig:translation_qual} shows both tasks on one subject; the tumor crops and error maps show where the methods differ.
LSF achieves the best mean scores across both translation tasks (Table~\ref{tab:main}). Relative to the strongest baseline for each task, change-region MAE decreases by 3.8\% for T1n$\to$T1c and 1.9\% for T2w$\to$T2f. Fig.~\ref{fig:translation_qual} shows the corresponding tumor crops and error maps. 

\noindent\textbf{Inpainting.} Table~\ref{tab:inpaint} compares LSF with three inpainting models on the test set ($n{=}125$). cWDM is the strongest baseline, with an average PSNR of 23.85\,dB and SSIM of 0.680. LSF improves on it by 1.29\,dB and 0.049 SSIM and is best in every contrast, with the largest gains on T2f (+2.1\,dB) and T2w (+2.1\,dB).

\subsection{Ablation}
\label{sec:exp_ablation}

Table~\ref{tab:ablation} builds LSF one component at a time: (b), the matched baseline, adds a local branch to (a); (c) sends $S$ to the global and $R$ to the local branch; 
(d) adds the input-token structural skip connection, yielding LSF.
(e) swaps the routing of (d); if the gain came only from splitting, (e) would match (d).
Each step improves every metric. The local branch adds 1.0\,dB over the global branch alone (a$\to$b). The split and routing (b$\to$c) change mainly where the error lies: $\rho_S$ drops from 11.5\% to 8.7\%, close to the 6.25\% of Gaussian error, while PSNR rises by 0.2\,dB. The structural skip (c$\to$d) changes mainly how large the error is, adding 0.6\,dB at a similar $\rho_S$. Reversing the routing (e) is worse than the baseline without the split, so the gain depends on the assignment, not on the split alone. cWDM, in pixel space, leaves 11.8\% of its error in the structure, the same share as the latent baseline (b).

\section{Conclusion}
\label{sec:conclusion}

%A pointwise latent loss measures how large an error is, not where it lies. In frozen 3D medical codecs, the damaging part of the error lies in a low-rank structure that is shared with the condition and correlated over long ranges, while the residual is local and specific to each image. LSF uses this split to route features and to predict the target, with the codec and the loss unchanged. It has the lowest error of all methods on two translation tasks and improves the best inpainting baseline by 1.3\,dB.
%LSF uses a content-dependent structure–residual decomposition to guide global/local processing and preserve input structure, while retaining a frozen codec and the pointwise flow objective. Experiments on MRI translation and tumor inpainting support this design, with a 1.29 dB inpainting PSNR gain over the strongest compared baseline.
%In frozen 3D medical codecs, attenuating the low-rank structure of a latent window distorts the decoded image more than an equal-norm change to the residual, and over-smoothed errors fall mostly in that structure. LSF routes the structure to a global branch and the residual to a local one and carries the input structure forward, with the codec and the flow loss unchanged. It achieves the best mean scores on two MRI translation tasks and gains 1.29\,dB PSNR over the strongest inpainting baseline; reversing the routing falls below the unsplit baseline.
We showed that the latent space of a frozen 3D medical codec is not flat: attenuating the low-rank structure of a latent window damages the decoded image more than a change of the same norm in the residual, and over-smoothed predictions put most of their error in the structure. LSF uses this split inside the generator: the structure goes to a global branch, the residual to a local branch, and the input structure is passed through directly. The codec and the loss stay the same. LSF gives the best results on two MRI translation tasks and improves tumor inpainting by 1.29\,dB PSNR over cWDM. Swapping the two branches is worse than not splitting at all.

% 06_prior_work merged into the introduction (v4)

% References should be produced using the bibtex program from suitable
% BiBTeX files. The IEEEbib.bst bibliography style file from IEEE produces
% unsorted bibliography list.
% -------------------------------------------------------------------------
% 参考文献用 \small 排版以收回溢出的一行（ICASSP 模板允许）
{\footnotesize
\bibliographystyle{IEEEbib}
\bibliography{strings,refs}
}

\end{document}